\documentclass{article} 
\usepackage{iclr2024_conference,times}

\usepackage{amsmath,amsfonts,bm}

\def\eqref#1{equation~\ref{#1}}

\def\1{\bm{1}}

\DeclareMathAlphabet{\mathsfit}{\encodingdefault}{\sfdefault}{m}{sl}
\SetMathAlphabet{\mathsfit}{bold}{\encodingdefault}{\sfdefault}{bx}{n}

\usepackage{hyperref}
\usepackage{url}
\usepackage{graphicx}
\usepackage{multirow}
\usepackage{multicol}
\usepackage{wrapfig}
\usepackage{caption}
\usepackage{subcaption}
\usepackage{booktabs}

\title{PyGS: Large-scale Scene Representation with Pyramidal 3D Gaussian Splatting}

\author{Zipeng Wang \& Dan Xu \\
Department of Computer Science and Engineering \\
The Hong Kong University of Science and Technology (HKUST) \\
Clear Water Bay, Kowloon, Hong Kong \\
zwang253@connect.hkust-gz.edu.cn, danxu@cse.ust.hk
}

\iclrfinalcopy 
\begin{document}

\maketitle

\begin{abstract}
Neural Radiance Fields (NeRFs) have demonstrated remarkable proficiency in synthesizing photorealistic images of large-scale scenes. 
However, they are often plagued by a loss of fine details and long rendering durations.
3D Gaussian Splatting has recently been introduced as a potent alternative, achieving both high-fidelity visual results and accelerated rendering performance. 
Nonetheless, scaling 3D Gaussian Splatting is fraught with challenges.
Specifically, large-scale scenes grapples with the integration of objects across multiple scales and disparate viewpoints, which often leads to compromised efficacy as the Gaussians need to balance between detail levels. 
Furthermore, the generation of initialization points via COLMAP from large-scale dataset is both computationally demanding and prone to incomplete reconstructions.
To address these challenges, we present \textbf{Py}ramidal 3D \textbf{G}aussian \textbf{S}platting (\textbf{PyGS}) with NeRF Initialization. 
Our approach represent the scene with a hierarchical assembly of Gaussians arranged in a pyramidal fashion. 
The top level of the pyramid is composed of a few large Gaussians, while each subsequent layer accommodates a denser collection of smaller Gaussians. 
We effectively initialize these pyramidal Gaussians through sampling a rapidly trained grid-based NeRF at various frequencies.
We group these pyramidal Gaussians into clusters and use a compact weighting network to dynamically determine the influence of each pyramid level of each cluster considering camera viewpoint during rendering.
Our method achieves a significant performance leap across multiple large-scale datasets and attains a rendering time that is over 400 times faster than current state-of-the-art approaches.
Our project page is available at \url{https://wzpscott.github.io/pyramid_gaussian_splatting/}.

\end{abstract}

\section{Introduction}

\begin{figure*}[t]
    \centering
    \includegraphics[width=1.\linewidth]{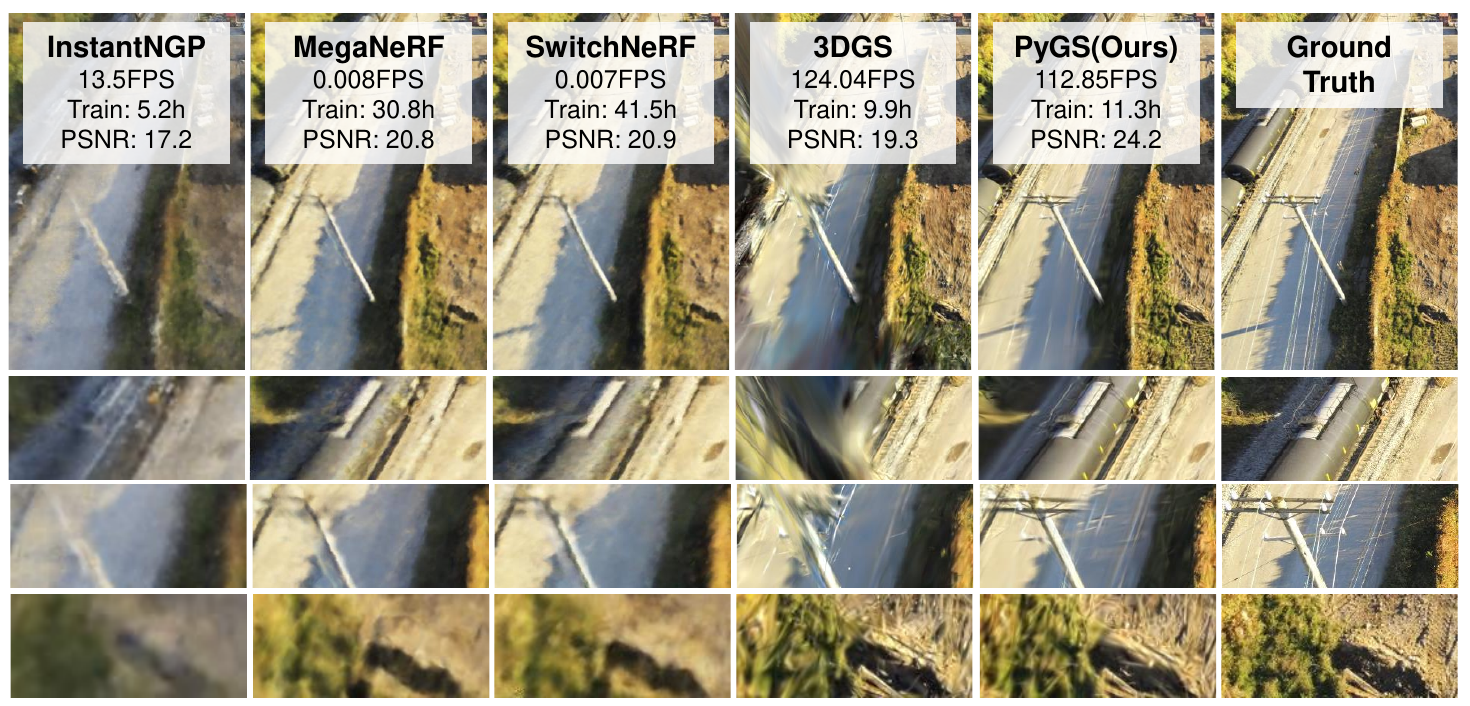}
    \caption{
    We significantly enhance the capability of 3D Gaussian Splatting (3DGS)~\cite{kerbl20233dgs} to model large-scale scenes while maintaining comparable training and rendering costs.
    The key is a pyramidal Gaussian structure and dynamic weighting of each level's contribution through a gating network. 
    Our method captures extremely high-frequency details and achieves a rendering speedup of over 400 times compared to state-of-the-art NeRF-based methods~\cite{turki2022meganerf, zhenxing2023switchnerf}.
    }
    \label{fig:coverfig}
\end{figure*}

Neural Radiance Fields (NeRF)~\cite{mildenhall2021nerf} have attracted much attention for 3D scene representation. 
Several studies, including MegaNeRF~\cite{turki2022meganerf}, BlockNeRF~\cite{tancik2022block}, and SwitchNeRF~\cite{zhenxing2023switchnerf}, have endeavored to extend NeRF to accommodate larger scenes, such as buildings or entire cityscapes. 
While these methods offer substantial enhancements over the original NeRF~\cite{mildenhall2021nerf},
they still tend to favor the learning of the scene's low-frequency elements due to spectral biases~\cite{rahaman2019spectral, tancik2020fourier} of neural networks, which may result in an inability to capture intricate details.
Moreover, the rendering process is exceedingly slow, mainly due to the heavy computational demands associated with the extensive sampling required for volumetric rendering.
These challenges significantly hinder their practicality for real-time applications.
Recently, 3D Gaussian Splatting (3DGS)~\cite{kerbl20233dgs} has emerged as a notable leap forward in the field of 3D scene modeling, acclaimed for its ability to render high-quality images of high resolution in real time. 
3DGS~\cite{kerbl20233dgs} represents a scene within a collection of 3D Gaussians, each endowed with learnable color and shape parameters. 
3DGS~\cite{kerbl20233dgs} excels at modeling exceedingly high-frequency details while also ensuring that the model size and rendering time grow linearly that aligns with the scene's size. 
These properties position 3DGS as a compelling alternative to NeRF-based methods for representing of large-scale scenes.

Nevertheless, the extension of 3D Gaussian Splatting (3DGS) to large-scale scenes, such as urban environments, presents a multitude of challenges. 
While the initial 3DGS framework~\cite{kerbl20233dgs} is optimized for small outdoor environments, where the employment of a single-scale Gaussian model is adequate. However, this approach falls short in more extensive scenes.
Large-scale scenes are characterized by a diverse collection of objects of widely varying sizes and detail levels. 
For example, imposing structures like skyscrapers or bridges may coexist with smaller features such as street signs and vegetation.
The conventional 3DGS method models the scene with a same set of 3D Gaussians for every viewpoint. This approach is not well-equipped to handle the varying scales inherent in large-scale scenes and often falls short in representing all the different scales accurately.
Additionally, 3DGS~\cite{kerbl20233dgs} depends on COLMAP~\cite{schoenberger2016mvs, schoenberger2016sfm} to generate sparse point clouds for initializing the algorithm. This strategy is generally effective for datasets with a limited size, typically encompassing hundreds of images. However, the majority of existing large-scale datasets~\cite{turki2022meganerf, UrbanScene3D, xiangli2022bungeenerf, li2023matrixcity} consist of thousands of images, capturing spaces that are orders of magnitudes larger. 
The substantial increase in the volume of data can lead to significant time increase in computing initial point clouds via COLMAP.
Furthermore, the resulting point clouds often lack representations of background elements, such as the sky and distant buildings, due to the difficulty in estimating their depth.

To overcome these challenges, we introduce Pyramidal Gaussian Splatting (PyGS), a novel multi-scale 3D Gaussian Splatting framework for large-scale scenes. 
PyGS parameterizes the scene using a hierarchical structure of 3D Gaussians, organized into pyramid levels that represent different levels of details.
The top level of the pyramid comprises a few large Gaussians that encapsulate general scene shape and structure, and the lower levels contain more smaller Gaussians that model finer scene details.
To initiate the pyramidal Gaussians, we utilize a coarse grid-based NeRF~\cite{muller2022instant}, which enables us to \emph{rapidly} create an initial and rough 3D model of the scene. 
In contrast to COLMAP, which exhibits a marked increase in processing time with larger datasets, grid-based NeRF approaches~\cite{chen2022tensorf, muller2022instant} demonstrate a performance that is largely unaffected by the size of the training set.
We subsequently sample the grid-based NeRF to form a dense point cloud and generate several subsets of point clouds by sampling the dense one at multiple frequencies.
These subsets of points serve as the initialization points for the respective levels of the proposed pyramidal Gaussian structure.

When rendering an image, it is desirable to adaptively choose the contribution of each level based both on the camera viewpoint and the complexity of the rendered region.
For example, when the camera is distant or the region is smooth and textureless, the higher-level Gaussians (i.e., fewer and larger Gaussians) should be prioritized. Conversely, when the camera is in close proximity or the region has complex geometry or textures, the lower levels (i.e., more smaller Gaussians) should come into play. 
To achieve this, we design a compact weighting network that dynamically determines the weights of each pyramid level.
We group the pyramidal Gaussians into clusters with the mini-batched K-means algorithm and use shared weights within each cluster.
To more effectively capture the nuances of local geometry and texture, we introduce a learnable embedding for each cluster. 
This embedding, along with the camera viewpoint, is input into the weighting network, which then computes the weights for the different pyramid levels associated with that cluster. 
Consequently, Gaussians belonging to the same level and cluster are assigned the same weight.
Additionally, we integrate an appearance embedding for each cluster and a color correction network to account for the intricate variations in lighting that can occur across viewpoints.

In this paper, we present the following contributions:
\textbf{(i)} We pioneer the application of 3D Gaussian Splatting (3DGS)~\cite{kerbl20233dgs} to large-scale scene modeling, addressing the challenges of preserving fine details and reducing the rendering times.
\textbf{(ii)} We introduce a novel pyramidal Gaussian splatting framework that effectively captures multi-scale information and adaptively renders different levels of scene details based on a specially designed weighting network.
\textbf{(iii)} We develop an efficient initialization technique utilizing a coarsely trained NeRF, which significantly shortens the preprocessing time duration and enhances the performance.
\textbf{(iv)} Through extensive experimentation across four large-scale datasets, we demonstrate the superior performance and efficiency of our proposed method.

\section{Related Work}
\subsection{Large-scale Scene Representations}
Learning digital representations of large-scale scenes, such as buildings and street blocks, has been a persistent challenge within the computer vision community. Early efforts~\cite{agarwal2011buildingrome, frahm2010buildingrome2, fruh2004automated, pollefeys2008detailed, snavely2006phototourism} focused on developing robust and parallel systems capable of reconstructing dense 3D geometry from large image collections. These methods generally depend on feature matching~\cite{lowe2004sift} and bundle adjustment~\cite{agarwal2010bundle1, triggs2000bundle2} to generate the corresponding camera poses and sparse point clouds for each image. 
To obtain denser representations, the sparse output can be refined using dense multiview stereo techniques\cite{goesele2007mvs}. However, these approaches may encounter artifacts and limited texture quality due to scaling issues~\cite{tancik2022block}.

Recently, neural radiance fields~(NeRF)~\cite{mildenhall2021nerf} have demonstrated significant performance improvements and flexibility in parameterizing 3D scenes. However, the standard NeRF may struggle to accurately represent extensive scenes due to limited model capacity. Several studies~\cite{turki2022meganerf, tancik2022block, zhenxing2023switchnerf} have tackled this issue using a divide-and-conquer approach, which involves decomposing the scene into multiple parts and training individual NeRF networks for each part. 
Although these decomposition-based techniques yield better results compared to standard NeRF, they generally require significantly more computational resources to capture intricate details in large scenes. For example, Mega-NeRF~\cite{turki2022meganerf} employs eight sub-networks, each comprising two NeRF networks, to represent a single scene. This leads to a complex architecture and increased training and rendering times.

\subsection{Differentiable Point-based Rendering}
Differentiable point-based rendering techniques~\cite{xu2022pointnerf, kerbl20233dgs} model scenes using discrete, unstructured point clouds. 
Specifically, Point-NeRF~\cite{xu2022pointnerf} parameterizes scenes into point clouds with neural features, allowing for interpolation between neighboring points to determine the value at any spatial coordinate. This approach, which uses the same volumetric rendering pipeline as NeRF~\cite{mildenhall2021nerf}, has been shown to yield superior reconstruction quality and reduced training times.

In a more recent development, 3D Gaussian splatting~\cite{kerbl20233dgs} has been introduced as a strong contender to NeRF, offering a scene parameterization through anisotropic Gaussians. 3DGS achieves image rendering by efficiently sorting and projecting these Gaussians onto image planes. Given that the computational effort for sorting and projection is fixed irrespective of the rendered resolution, 3DGS demonstrates a significant advantage in rendering speed, particularly for images of high resolution.
Despite the impressive performance of 3DGS~\cite{kerbl20233dgs}, it faces challenges with capturing varying levels of details in expansive scenes and the time-intensive process of initialization using COLMAP~\cite{schoenberger2016mvs, schoenberger2016sfm} for large-scale datasets. 
Our approach address those problems by introducing a novel pyramidal Gaussian architecture that dynamically adjusts level weights through a weighting network, coupled by extracting initial point clouds from a coarsely trained grid-based NeRF~\cite{muller2022instant, chen2022tensorf}.

\subsection{Multi-scale Neural Representations}
The concept of multi-scale representations~\cite{adelson1984pyramidmethods, burt1987laplacian}, has been a cornerstone for analyzing and processing images across different levels of resolutions, which has inspired the creation of multi-scale neural representations.
BACON~\cite{lindell2022bacon} employs band-limited constraints to guide the intermediate outputs at each layer, yielding a neural representation that progresses from coarse to fine in an interpretable manner. MINER~\cite{saragadam2022miner} constructs a Laplacian pyramid of neural fields which isolates portions of the signal across scales, fostering both sparsity and swift model convergence. 
In a similar vein, Bungee-NeRF~\cite{xiangli2022bungeenerf} introduces a model expansion strategy that incrementally introduces network layers to refine detail capture. Meanwhile, Instant-NGP~\cite{muller2022instant} adopts a multi-resolution hash grid that significantly enhances both training and inference speeds.
Despite these advances, a multi-scale architectural approach specifically tailored for discrete and unstructured 3D Gaussians remains unexplored in the field.

\subsection{Anti-aliasing Radiance Fields}
Aliasing presents a notable challenge for radiance fields, particularly when rendering images from various viewpoints. 
MipNeRF~\cite{barron2021mip} addresses this issue by sampling NeRF with conical frustum volumes rather than infinitesimal points, incorporating an integrated positional encoding to alleviate aliasing.
Further advancements for grid-based NeRF have also been made to efficiently counteract aliasing. 
PyNeRF~\cite{turki2024pynerf} trains multiple radiance fields for different spatial resolutions and selects the most suitable field based on the projected sample area. ZipNeRF~\cite{barron2023zipnerf} utilizes a multi-sampling strategy to approximate the average radiance over a conical frustum.
Tri-MipNeRF~\cite{hu2023trimip} and MipGrid~\cite{nam2024mip} address the problem using pre-filtering.
Recent developments have built upon 3DGS~\cite{kerbl20233dgs}. MipSplatting~\cite{yu2023mipsplatting} combines a 3D smoothing filter with a 2D Mip filter to enhance anti-aliasing capabilities.
In addition, Multi-scale Gaussian Splatting~\cite{yan2023multiscale3dgs} adjusts the scale of Gaussian filters in the rendering process, merging smaller Gaussians into larger ones for distant views to effectively manage aliasing at various distances and scales.


\section{Method}
\begin{figure*}[t!]
    \centering
    \includegraphics[width=0.99\linewidth]{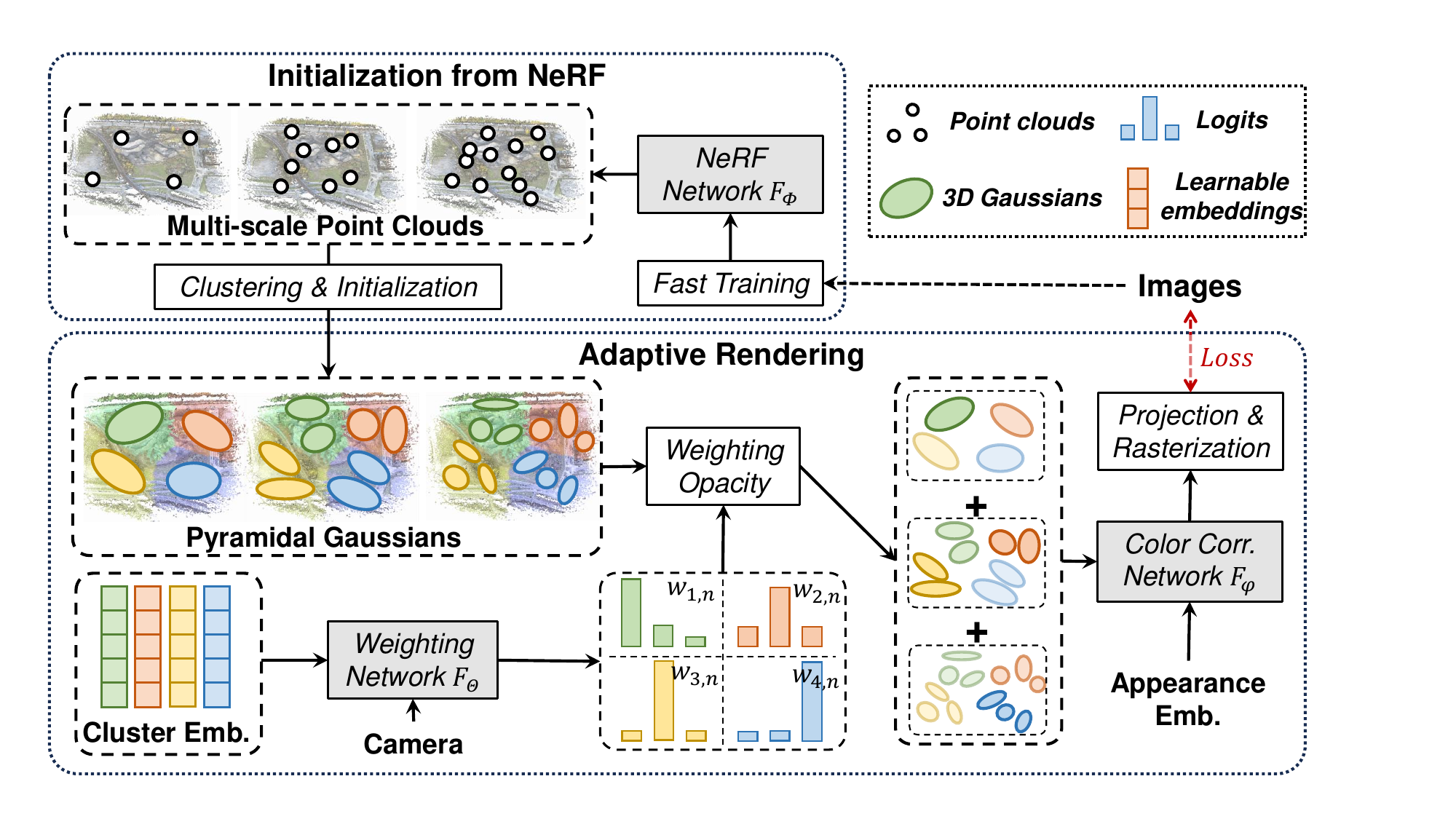}
    \caption{
    \textbf{Overview of the PyGS Framework.}
    Our framework begins by training a grid-based NeRF model to rapidly converge and generate point clouds with varying densities. 
    We then use the point clouds from NeRF to initialize a multi-level structure of Pyramidal Gaussians and spatially group these Gaussians into clusters.
    When rendering an image, a compact weighting network computes the contributions of different levels of each cluster, taking into account both the camera position and a learnable cluster embedding.
    Before these Gaussians are projected onto the image plane, a color correction network is applied to fine-tune the colors of the Gaussians to accommodate complex light changes.
    For clarity, the illustration shows 3 levels and 4 clusters.
    }
    \label{fig:framework}
\end{figure*}

\subsection{Preliminaries}

\textbf{3D Gaussian Splatting (3DGS)} represents scenes as a collection of 3D Gaussians.
Each Gaussian is characterized by its position, rotation, scale, opacity, and color (represented by a series of spherical harmonics). 
We denote the position, opacity, and color of the $i$-th Gaussian by $x_i$, $\alpha_i$, and $c_i$, respectively.

To render an image from a given viewpoint, the Gaussians are projected onto the image plane. They are then sorted in accordance with their depth values in the Z-buffer. The color $\hat{C}$ rendered for a pixel is determined by blending all Gaussians that are projected onto it in order, using the following equation:
\begin{equation}
    \hat{C} = \sum_{i=1}^{N} c_i \alpha_i \prod_{j=1}^{i-1} (1 - \alpha_j).
\end{equation}

\noindent \textbf{Neural Radiance Fields (NeRF) and Its Derivatives} represent the scene using an implicit function $F_\Phi$ that is typically parameterized by a multilayer perceptron (MLP):
\begin{equation}
    F_\Phi(\mathbf{x}, \mathbf{d}) = (c, \sigma),
\end{equation}
where $\mathbf{x}$ is the spatial coordinate, $\mathbf{d}$ is the view direction, and $c$ and $\sigma$ are the predicted color and density, respectively.
NeRF-based methods~\cite{mildenhall2021nerf, muller2022instant} render images by casting rays $r(t)$ from the camera into the scene, sampling points along the rays, and aggregating the samples via volume rendering techniques.
Certain NeRF-based methods~\cite{muller2022instant, chen2022tensorf} achieve extremely fast convergence by substituting the MLP with efficient data structures such as hash maps or tensor grids.

\subsection{Initialization of Pyramidal Gaussian from NeRF} \label{initialization}
\noindent \textbf{Point Cloud Generation from NeRF:}
The original 3D Gaussian Splatting (3DGS) method~\cite{kerbl20233dgs} utilizes COLMAP~\cite{schoenberger2016mvs, schoenberger2016sfm} to generate sparse point clouds as an initial step for the algorithm. 
However, large-scale datasets~\cite{turki2022meganerf, UrbanScene3D, xiangli2022bungeenerf, li2023matrixcity} are often more than an order of magnitude larger than typical outdoor datasets. The time required for sparse reconstruction with COLMAP grows substantially with the size of the dataset. It has also been observed that COLMAP struggles to reconstruct points that are distant from the scene's center, which critically impairs 3DGS's capacity to model far-away objects.
In contrast, grid-based NeRF methods~\cite{chen2022tensorf, muller2022instant} offer a more efficient and robust approach for point cloud generation that is largely independent of the size of the training set.

Motivated by this insight, we first train a grid-based NeRF model, denoted as $F_\Phi$, on the large-scale scene. 
As we only require the NeRF model to capture the scene's approximate structure, this training process can be expedited and completed within approximately 20 minutes. 
To generate point clouds from the trained NeRF model $F_\Phi$, we cast rays from the training cameras and sample points at the estimated termination point, denoted as $\hat{z}$, of each ray.
Please refer to the supplementary materials for a detailed derivation of the estimated terminations.
The position and color of each generated point are obtained as:
\begin{align}
    x_i &= \mathbf{o} + \hat{z} \mathbf{d}, \\
    c_i &= F_\Phi(x_i, \mathbf{d}),
\end{align}
where $\mathbf{o}$ and $\mathbf{d}$ represent the origin and direction of the ray, respectively.
We sample $N_\text{P}$ points from the coarse NeRF to form the point cloud, denoted by $P = \{(x_i, c_i) \mid 1 \le i \le N_\text{P}\}$.

\noindent \textbf{Sampling Multi-scale Point Clouds}
We sample $L$ subsets of points from $P$ with progressively increasing sampling frequencies.
Specifically, for the $l$-th subset of points, we randomly select $N_l$ points from $P$. 
We empirically set $N_l = l \cdot N_1$, where $N_1$ denotes the number of points in the first and most sparse subset of points. 
We denote $N = \sum_{l=1}^L N_i$ as the total number of selected points.

\noindent \textbf{Initiating Pyramidal Gaussians:}
For each subset of points, we initialize 3D Gaussians using the same approach in 3DGS~\cite{kerbl20233dgs}, details of which can be found in the supplementary materials.
This leads to a hierarchical structure of 3D Gaussian collections, which we refer to as \textbf{Pyramidal Gaussians}, denoted by $G_{\text{py}} = {G_l \mid 1 \le l \le L}$, where $G_l$ represents the Gaussians initiated from the $l$-th sub-cloud.
Since the scale of each Gaussian is determined by the average distance to its nearest neighbors, the scale of the Gaussians diminishes as the number of initiated points increases. Consequently, the Pyramidal Gaussians $G_{\text{py}}$ naturally form a multi-scale architecture. The higher levels (i.e., smaller $l$) comprise Gaussians with larger scales, while the lower levels consist of Gaussians with smaller scales.

\subsection{Adaptive Rendering}
\begin{wrapfigure}{l}{.5\textwidth}
\centering
\includegraphics[width=0.9\linewidth]{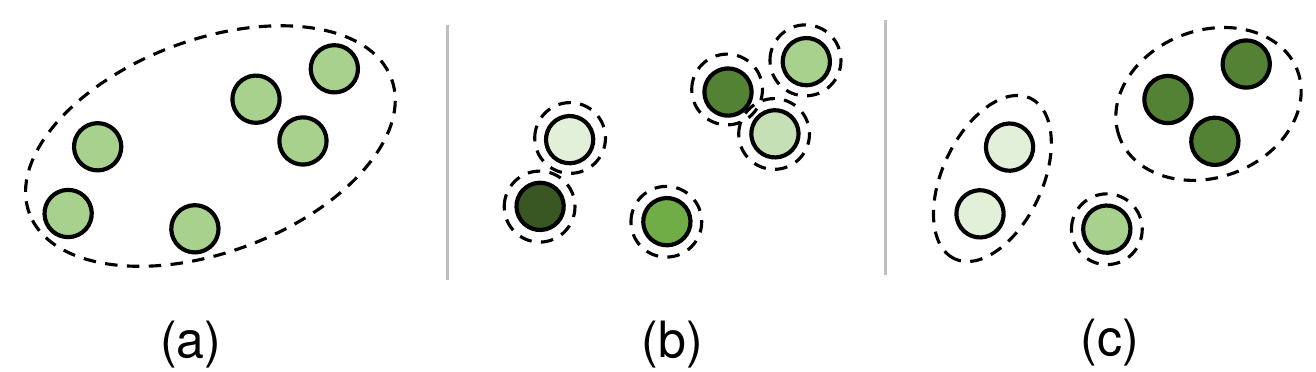}
    \caption{
    Illustration of weighting strategies.}
    \label{fig:weighting-strategy}
\end{wrapfigure}

\noindent \textbf{Cluster-level Weighting:}
When rendering the Pyramidal Gaussians $G_\text{py}$, it is crucial to assess the contributions from each level of Gaussians adaptively based on the viewpoint. 
Fig.~\ref{fig:weighting-strategy} illustrates three different weighting strategies: (a) uniform weighting, which assigns the same weight to all Gaussians within the same level; (b) unique weighting, which assigns a distinct weight to each Gaussian; and (c) cluster weighting, which groups Gaussians into clusters and assigns a shared weight to all members within each cluster.

\noindent Uniformly weighting all Gaussians from the same level may not be ideal, as the appropriate weight should reflect specific characteristics of each Gaussian, such as its distance to the camera and the surrounding geometric complexity. Assigning a unique weight to each Gaussian could provide a more tailored rendering, but would also incur a substantial computational cost due to the large number of Gaussians involved. 
We therefore choose a balanced scheme that segments the Pyramidal Gaussians into clusters and attributes each cluster with an unique set of weights. This strategy enables us to consider the positional information of the Gaussians while avoiding a significant increase in computational demand.

We employ an efficient mini-batched K-means algorithm~\cite{shindler2011fast} to establish $K$ centroids from $P$, denoted as $M=\{m_k \mid 1 \le k \le K\}$, where $m_k$ represents the $k$-th centroid. 
Note that to decrease computational time, we apply the K-means algorithm to only a subset of the point cloud. The duration of the K-means algorithm is approximately 30 seconds.
For every Gaussian in $G_\text{py}$, we assign it to the nearest centroid from the set $M$. For the $i$-th Gaussian in $G_l$, this assignment is represented by:
\begin{equation}
    k_i^l = \text{NearestNeighbour}(M, x_i^l),
\end{equation}
where $k_i^l$ is the index of the closest centroid to the ellipsoid center of the $i$-th Gaussian in $G_l$, as determined by the Nearest Neighbour algorithm.
Here Gaussians from different levels that belong to the same cluster also make up a pyramidal architecture that captures a small region near the cluster centroid.

\label{eq:assign_cluster} 
\begin{wrapfigure}{r}{.25\textwidth}
\centering
\includegraphics[width=0.95\linewidth]{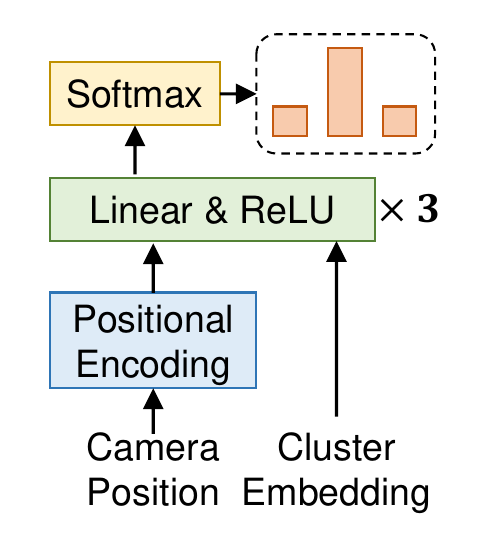}
    \caption{
    Architecture of weighting network.
    }
    \label{fig:arch}
\end{wrapfigure}

\noindent \textbf{Cluster Embedding:}
Various attributes of clusters, such as their position, local geometry, and texture, can influence the optimal weights for each level.
To better capture these cluster information, we equip each cluster with a learnable feature, referred to as a cluster embedding $E_\text{c} \in \mathbb{R}^{K \times D}$, where $K$ represents the total number of clusters and $D$ is the dimension of the feature. 

\noindent \textbf{Weighting Network:}
We employs a small MLP $F_\Theta$ to determine the level weights for each cluster.
The architecture of the weighting network is visualized in Fig~\ref{fig:arch}.
For the $k$-th cluster as viewed from the $n$-th camera, the level weights are computed as follows:
\begin{equation}
    w_{k,n} = \text{Softmax}\left(
        F_\Theta\left(
            \text{Concat}\left(E_\text{c}[k], \text{PE}(\mathbf{x}_\text{cam})\right)
        \right)
    \right) \in \mathbb{R}^L,
\end{equation}
where $\text{Concat}(\cdot)$ denotes the concatenation operation, $E_\text{c}[k]$ is the feature vector for the $k$-th cluster, $\text{PE}$ represents a positional encoding function~\cite{mildenhall2021nerf}, and $\mathbf{x}_\text{cam}$ denotes the coordinates of the camera center. 
The weight corresponding to the $l$-th level for the $k$-th cluster viewed from the $n$-th camera is denoted by $w_{k,n}[l]$. 

\noindent \textbf{Appearance-aware Rendering:}
To render an image, we sort all the Gaussians using the same method as presented in 3DGS~\cite{kerbl20233dgs}. 
With a slight abuse of notations, we denote the sorted Gaussians as a sequence $\{g_i\}_{i=1}^N$, where $g_i$ is the $i$-th ordered Gaussian.
The color, opacity, level, and cluster index associated with $g_i$ are represented by $c_i$, $\alpha_i$, $l_i$, and $k_i$, respectively.

As large-scale scene modeling frequently encounters challenges stemming from lighting variations among different images, prior works such as~\cite{turki2022meganerf, zhenxing2023switchnerf} have considered appearance embedding vectors to impart additional adaptability for handling these variations.
However, directly incorporating such a strategy into our framework may not be feasible, we have devised an effective alternative. 
We integrate a per-camera appearance embedding $E_\text{app} \in \mathbb{R}^{N_\text{cam} \times D}$ alongside a compact MLP $F_\varphi$ to perform per-cluster color correction. Here, $N_\text{cam}$ represents the total count of cameras.
The color correction for the $i$-th Gaussian is computed as follows: \begin{equation} 
\Delta c_i = F_\varphi(\text{Concat}(E_\text{c}[k_i], E_\text{app}[n])) \in \mathbb{R}^3, 
\end{equation} 
where $E_\text{c}[k_i]$ denotes the embedding of the cluster to which the $i$-th Gaussian belongs, and $E_\text{app}[n]$ is the appearance embedding for the $n$-th camera.
The final color for rendering is then determined by the following equation: \begin{equation} 
\hat{C} = \sum_{i=1}^{N} (c_i + \Delta c_i) \cdot w_{k_i, n}[l_i] \cdot \alpha_i \prod_{j=1}^{i-1} (1 - w_{k_j, n}[l_j] \cdot \alpha_j), 
\end{equation} 
where $c_i$ is the color computed via spherical harmonics, $c_i' = c_i + \Delta c_i$ is the adjusted color based on the current image, and $\alpha_i$ is the opacity of the Gaussian.

\subsection{Optimization}
Similar to the original 3DGS~\cite{kerbl20233dgs}, we use a photorealistic loss function for optimization, which is a weighted combination of mean squared error (MSE) and structural similarity index measure (SSIM):
\begin{equation}
    \mathcal{L} = \lambda \cdot \text{MSE}(C, \hat{C}) + (1 - \lambda) \cdot \text{SSIM}(C, \hat{C}),
\end{equation}
where $\lambda$ is a hyperparameter that balances the two loss terms, and $C$ is the ground-truth color of the pixel.
We use the same splitting and cloning strategies as 3DGS for density control. 
The newly split and cloned Gaussians will be retained within the original level and cluster.
In order to preserve the multi-scale characteristics throughout the training process, we routinely reassess and reassign the levels of Gaussians in accordance with their scales. This ensures that the distribution of Gaussians across different levels remains proportionally balanced.

\section{Experiments}
\begin{table}[!h]
\centering
\caption{The testing results of our PyGS on four large-scale datasets~\cite{turki2022meganerf, UrbanScene3D, li2023matrixcity, xiangli2022bungeenerf} Our method gets state-of-the-art accuracy and significant rendering speed boost compared with NeRF-based methods~\cite{turki2022meganerf, zhenxing2023switchnerf, xiangli2022bungeenerf, xu2023gridnerf} and original 3DGS~\cite{kerbl20233dgs}.}
    \begin{subtable}[h]{1\columnwidth}
    \centering
    \resizebox{0.95\columnwidth}{!}{%
    \begin{tabular}{@{}c|c|cccccc|c@{}}
    \toprule
    Scene                                                                           & Metric & NeRF  & NeRF++ & MegaNeRF & SwitchNeRF & 3D-GS(R) & 3D-GS(C) & PyGS   \\ \midrule
    \multirow{3}{*}{\begin{tabular}[c]{@{}c@{}}Mill19\\ Building\end{tabular}}      & PSNR$\uparrow$   & 19.54 & 19.48  & 20.93    & \underline{21.54}      & 19.09    &20.28& \textbf{22.72}  \\
                                                                                    & SSIM$\uparrow$   & 0.525 & 0.520  & 0.547    & 0.579      & 0.562    &\underline{0.647}& \textbf{0.729}\\
                                                                                    & LPIPS$\downarrow$  & 0.512 & 0.514  & 0.504    & 0.474      & 0.468    &\underline{0.365}& \textbf{0.298}  \\
                                                                                    \midrule
    \multirow{3}{*}{\begin{tabular}[c]{@{}c@{}}Mill19\\ Rubble\end{tabular}}        & PSNR$\uparrow$   & 21.14 & 20.90  & 24.06    & \underline{24.31}      & 23.39    & 24.75& \textbf{25.65}  \\
                                                                                    & SSIM$\uparrow$   & 0.522 & 0.519  & 0.553    & 0.562      & 0.610&\underline{0.676}& \textbf{0.727}\\
                                                                                    & LPIPS$\downarrow$  & 0.546 & 0.548  & 0.516    & 0.496      & \underline{0.467}    &          0.343& \textbf{0.320}\\
                                                                                    \midrule
    \multirow{3}{*}{\begin{tabular}[c]{@{}c@{}}Urbanscene\\ Residence\end{tabular}} & PSNR$\uparrow$   & 19.01 & 18.99  & 22.08    & \underline{22.57}      & 20.42    &21.13& \textbf{23.74}  \\
                                                                                    & SSIM$\uparrow$   & 0.593 & 0.586  & 0.628    & 0.654      & 0.689    &\underline{0.744}& \textbf{0.789}  \\
                                                                                    & LPIPS$\downarrow$  & 0.488 & 0.493  & 0.489    & 0.457      & 0.342    &\underline{0.272}& \textbf{0.260}  \\
                                                                                    \midrule
    \multirow{3}{*}{\begin{tabular}[c]{@{}c@{}}Urbanscene\\ Sci-Art\end{tabular}}   & PSNR$\uparrow$   & 20.70 & 20.83  & 25.60    & \underline{26.52}      & 21.41    &22.11& \textbf{27.82}  \\
                                                                                    & SSIM$\uparrow$   & 0.727 & 0.755  & 0.770    & 0.795      & 0.757    &\underline{0.818}& \textbf{0.864}  \\
                                                                                    & LPIPS$\downarrow$  & 0.418 & 0.393  & 0.390    & 0.360      & 0.349    &\underline{0.252}& \textbf{0.214}  \\
                                                                                    \midrule
    \multirow{3}{*}{\begin{tabular}[c]{@{}c@{}}Urbanscene\\ Campus\end{tabular}}    & PSNR$\uparrow$   & 21.83 & 21.81  & 23.42    & \underline{23.62}      & 14.06    &21.33& \textbf{24.25}  \\
                                                                                    & SSIM$\uparrow$   & 0.521 & 0.520  & 0.537    & 0.541      & 0.342    &\underline{0.642}& \textbf{0.745}  \\
                                                                                    & LPIPS$\downarrow$  & 0.630 & 0.630  & 0.618    & 0.609      & 0.745    &\underline{0.552}& \textbf{0.432}  \\
                                                                                    \bottomrule
    \end{tabular}%
    }
    \vspace{8pt}
    \caption{Results on Mill19~\cite{turki2022meganerf} and Urbanscene~\cite{UrbanScene3D} dataset.}
    \label{tab:mill19urbanscene}
    \end{subtable}
    \begin{subtable}[h]{1\columnwidth}
    \centering
    \resizebox{0.9\columnwidth}{!}{
    \begin{tabular}{@{}c|c|cccccc|c@{}}
    \toprule
    Scene                                                                         & Metric & NeRF& InstantNGP &  MipNeRF-360&GridNeRF & 3D-GS(R) & 3D-GS(C) & PyGS \\ \midrule
    \multirow{3}{*}{\begin{tabular}[c]{@{}c@{}}MatrixCity\\ Block A\end{tabular}} & PSNR$\uparrow$   &      23.15&            27.21&           26.64&25.37&          23.69& \underline{27.33}&      \textbf{27.96}\\
                                                                                  & SSIM$\uparrow$   &      0.561&            0.793&           0.772&0.705&          0.658& \underline{0.852}&      \textbf{0.863}\\
                                                                                  & LPIPS$\downarrow$  &      0.649&            0.376&           0.406&0.478&          0.514& \underline{0.225}&      \textbf{0.214}\\
                                                                                  \midrule
    \multirow{3}{*}{\begin{tabular}[c]{@{}c@{}}MatrixCity\\ Block C\end{tabular}} & PSNR$\uparrow$   &      22.15&            23.21&           24.20&23.65&          20.81& \underline{24.34}&      \textbf{24.77}\\
                                                                                  & SSIM$\uparrow$   &      0.590&            0.788&           0.759&0.741&          0.5986& \underline{0.807}&      \textbf{0.819}\\
                                                                                  & LPIPS$\downarrow$  &      0.527&            0.311&           0.365&0.389&          0.4299& \underline{0.214}&      \textbf{0.205}\\
                                                                                  \midrule
    \multirow{3}{*}{\begin{tabular}[c]{@{}c@{}}MatrixCity\\ Block E\end{tabular}} & PSNR$\uparrow$   &      23.53&            26.36&           26.54&24.43&          22.09& \underline{26.93}&      \textbf{27.62}\\
                                                                                  & SSIM$\uparrow$   &      0.612&            0.807&           0.811&0.693&          0.6065& \underline{0.8543}&      \textbf{0.872}\\
                                                                                  & LPIPS$\downarrow$  &      0.534&            0.335&           0.338&0.468&          0.4738& \underline{0.189}&      \textbf{0.181}\\
                                                                                  \bottomrule
    \end{tabular}%
    }
    \label{tab:matrixcity}
    \caption{Results on MatrixCity~\cite{li2023matrixcity} dataset.}
    \end{subtable}
    \begin{subtable}[h]{1\columnwidth}
    \centering
    \resizebox{0.85\columnwidth}{!}{
    \begin{tabular}{@{}c|c|ccccc|c@{}}
    \toprule
    Scene                                                                             & Metric & NeRF & MipNeRF & BungeeNeRF & 3DGS-R& 3DGS-C& PyGS \\ \midrule
    \multirow{3}{*}{\begin{tabular}[c]{@{}c@{}}BungeeNeRF\\ 56Leonard\end{tabular}}   & PSNR$\uparrow$   &      21.70&         22.312&            24.513&          26.4&          \underline{28.4}&      \textbf{29.22}\\
                                                                                      & SSIM$\uparrow$   &      0.636&         0.689&            0.815&          0.847&          \underline{0.904}&      \textbf{0.944}\\
                                                                                      & LPIPS$\downarrow$  &      0.320&         0.266&            0.139&          0.102&          \underline{0.088}&      \textbf{0.074}\\
                                                                                      \midrule
    \multirow{3}{*}{\begin{tabular}[c]{@{}c@{}}BungeeNeRF\\ Tranamerica\end{tabular}} & PSNR$\uparrow$   &      22.642&         22.692&            24.415&          24.22&          \underline{26.94}&      \textbf{27.79}\\
                                                                                      & SSIM$\uparrow$   &      0.690&         0.687&            0.801&          0.823&          \underline{0.896}&      \textbf{0.923}\\
                                                                                      & LPIPS$\downarrow$  &      0.318&         0.327&            0.192&          0.231&          \underline{0.148}&      \textbf{0.108}\\
                                                                                      \bottomrule
    \end{tabular}
    }
    \label{tab:bungeenerf}
    \caption{Results on BungeeNeRF~\cite{xiangli2022bungeenerf} dataset.}
    \end{subtable}
\label{tab:results}
\end{table}
\begin{figure}[!h]
    \centering
    \begin{subfigure}[h]{1\textwidth}
        \centering
        \includegraphics[width=.99\linewidth]{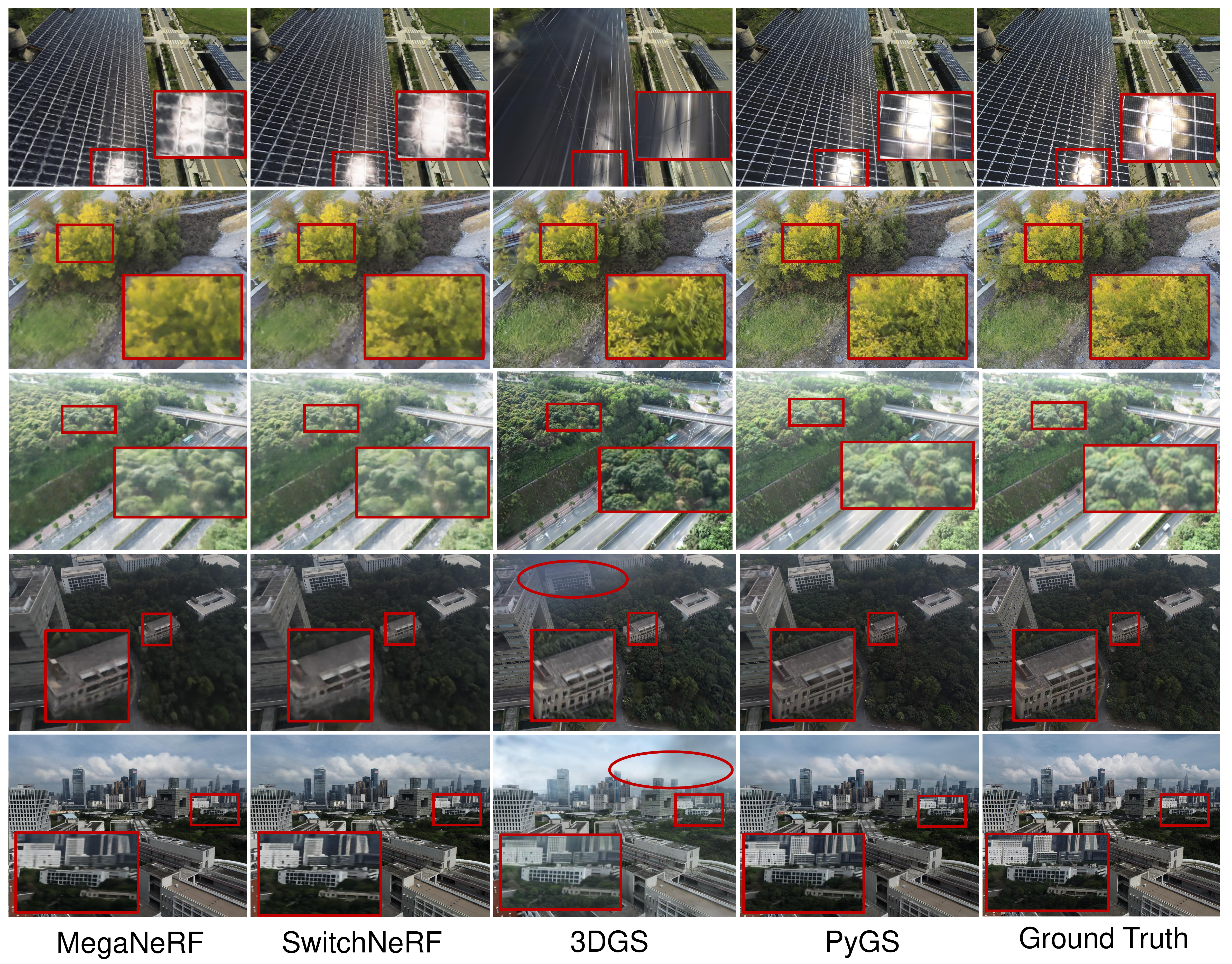}
        \caption{
        The comparison of the rendered images on Mill19~\cite{turki2022meganerf} and Urbanscene~\cite{UrbanScene3D} dataset.
        }
    \end{subfigure}
    \hfill
    \begin{subfigure}[h]{1\textwidth}
        \centering
        \includegraphics[width=.996\linewidth]{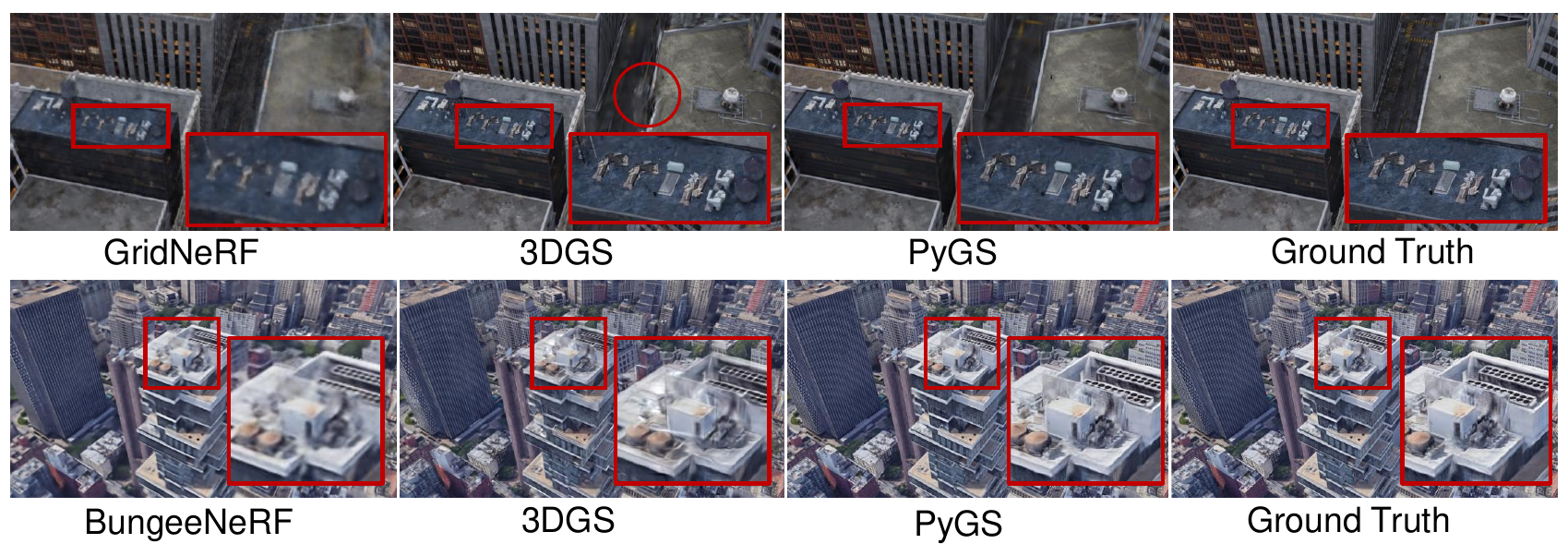}
        \caption{
        The comparison of the rendered images on BungeeNeRF~\cite{xiangli2022bungeenerf} and MatrixCity~\cite{li2023matrixcity} dataset. 
        }
    \end{subfigure}
    \caption{
        The comparison of the rendered images from Mega-NeRF~\cite{turki2022meganerf}, Switch-NeRF~\cite{zhenxing2023switchnerf}, GridNeRF~\cite{xu2023gridnerf}, BungeeNeRF~\cite{xiangli2022bungeenerf}, the original 3DGS~\cite{kerbl20233dgs} and our PyGS on four large-scale dataset. 
        Our method exhibits finer details and captures tiny structures more effectively than the compared methods. 
        Additionally, our approach offers improved background modeling and reduces the presence of floaters when compared to 3DGS. 
        Please zoom in to explore the details.
        }
    \vspace{-20pt}
\label{fig:comparison}
\end{figure}
\subsection{Experimental Setup}
\noindent \textbf{Dataset:}
We conduct our experiments on four large-scale scene datasets: Mill19~\cite{turki2022meganerf}, Urbanscene~\cite{UrbanScene3D}, MatrixCity~\cite{li2023matrixcity}, and BungeeNeRF~\cite{xiangli2022bungeenerf}. 
Mill19~\cite{turki2022meganerf} and Urbanscene~\cite{UrbanScene3D} collectively comprise five real-world large-scale urban scenes captured by drones. 
Additionally, we adapt three scenes from the MatrixCity~\cite{li2023matrixcity} dataset, which is a synthetic city-level dataset generated with Unreal Engine 5. 
Furthermore, our method is evaluated on two scenes from BungeeNeRF~\cite{xiangli2022bungeenerf}, sourced from Google Earth Studio~\cite{GoogleEarthStudio}. 
These scenes encompass a wide range of scales, from very distant to extremely close views.
We report main results on all the four chosen datasets and conduct analysis experiments on Mill19~\cite{turki2022meganerf} and Urbanscene~\cite{UrbanScene3D} datasets.

\noindent \textbf{Baselines:}
We evaluat our PyGS method against state-of-the-art (SoTA) methods on each dataset. 
For the Mill19 and Urbanscene datasets, we compare PyGS with NeRF~\cite{mildenhall2021nerf}, NeRF++~\cite{zhang2020nerf++}, and two SoTA NeRF-based approaches, namely MegaNeRF~\cite{turki2022meganerf} and SwitchNeRF~\cite{zhenxing2023switchnerf}. 
On the MatrixCity dataset, our comparisons include NeRF~\cite{mildenhall2021nerf}, InstantNGP~\cite{muller2022instant}, MipNeRF-360~\cite{barron2022mip360}, and GridNeRF~\cite{xu2023gridnerf}. 
For the BungeeNeRF dataset, we benchmark against NeRF~\cite{mildenhall2021nerf}, MipNeRF~\cite{barron2021mip}, and the dataset's namesake method, BungeeNeRF~\cite{xiangli2022bungeenerf}. Furthermore, we compare PyGS with the original 3DGS~\cite{kerbl20233dgs} using random points (3DGS-R) and COLMAP points (3DGS-C) for initialization on all the four datasets.

\noindent \textbf{Implementation:}
In our main experiments, we employ Pyramidal Gaussians with $L=3$ levels and $K=5000$ clusters. 
We use an instant-NGP~\cite{muller2022instant} to generate $10^6$ points for each scene.
The number of initiation points for each level is set to $8 \times 10^5$, $1.6 \times 10^6$, and $3.2 \times 10^6$, respectively. 
Both the weighting network and the color correction network are configured with 3 layers, each containing 64 neurons using fully-fused MLP~\cite{muller2022instant}.
The dimensions for the cluster and appearance embeddings are set to 64. The Adam optimizer~\cite{kingma2014adam} is employed, with learning rates set at $3 \times 10^{-4}$ for the networks and $0.1$ for the embeddings. 
We train model of each scene for 200,000 iterations on an Nvidia RTX A6000 GPU. 
The learning rate for the Gaussian positions is set to $1.6 \times 10^{-5}$, following the recommended practice for 3DGS in large scenes~\cite{kerbl20233dgs}. 
We maintain consistency with all other shared hyperparameters from the original 3DGS method~\cite{kerbl20233dgs}. 

\noindent \textbf{Evaluation Metrics:}
We quantitatively evaluate our results on novel view synthesis using PSNR and SSIM~\cite{wang2004ssim} (higher the better), as well as the VGG implementation of LPIPS~\cite{zhang2018unreasonable} (lower the better).

\subsection{Overall Results}
We present a qualitative comparison of the baseline methods and our approach in Fig.~\ref{fig:comparison}, as well as a quantitative analysis in Table~\ref{tab:results}. 
Our method outperforms the baseline methods across all evaluated metrics and datasets, achieving significantly higher scores. 
Our method exhibits a significant enhancement in rendering sharp edges and fine details within synthesized views, surpassing state-of-the-art NeRF techniques such as MegaNeRF~\cite{turki2022meganerf} and SwitchNeRF~\cite{zhenxing2023switchnerf}, which rely on scene partitioning strategies. 
Moreover, it outstrips the performance of GridNeRF~\cite{xu2023gridnerf} in tests conducted on the synthetic MatrixCity dataset~\cite{li2023matrixcity}.
Additionally, our approach adeptly addresses the challenges posed by the highly diverse scales within the BungeeNeRF dataset~\cite{xiangli2022bungeenerf}, yielding superior results compared to the current state-of-the-art method, BungeeNeRF itself. 
We have also noted that, in contrast to the original 3DGS method~\cite{kerbl20233dgs}, our approach proficiently reproduces intricate details, lighting effects, and background objects, which are aspects that 3DGS commonly struggles to capture accurately across all tested datasets.
Overall, the empirical results confirm that our method sets a new benchmark for quality and realism in the field of view synthesis.

\subsection{Model Analysis}
Besides the main results, we perform analysis experiments to analyze different aspects of our model on the Mill19~\cite{turki2022meganerf} dataset.

\begin{figure}[h!]
    \centering
    \includegraphics[width=.95\linewidth]{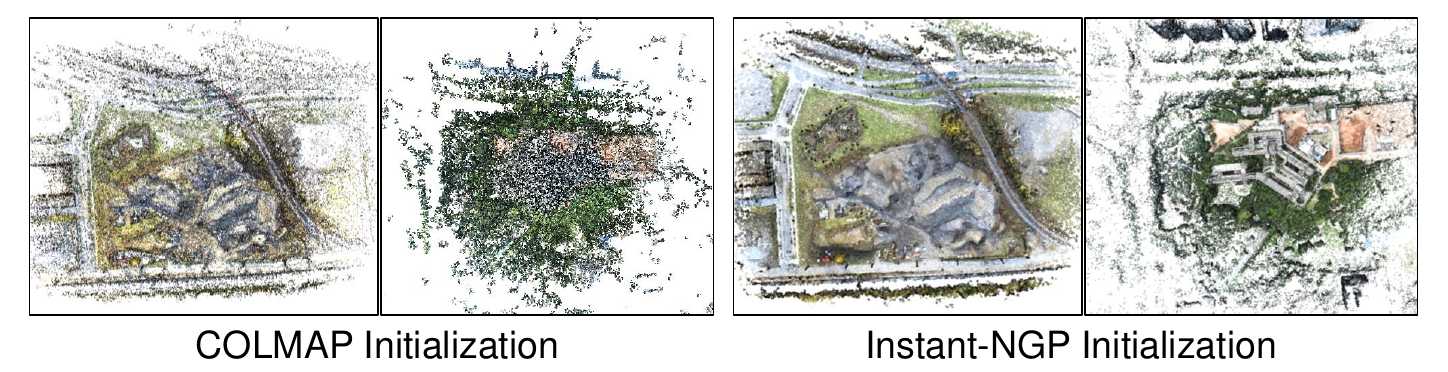}
    \caption{Comparison between COLMAP~\cite{schoenberger2016mvs, schoenberger2016sfm} and Instant-NGP~\cite{muller2022instant} point clouds. Instant-NGP produces point clouds with better modeling of texture and geometry.
    }
    \label{fig:pointclouds}
\end{figure}

\noindent \textbf{Effectiveness of NeRF-based Initialization:}
As shown in Table~\ref{tab:ablation}, we assess the performance of our NeRF-based initialization method by comparing it with two alternatives: random initialization and COLMAP-based initialization~\cite{schoenberger2016mvs, schoenberger2016sfm}. 
For the random approach, we generate a set of random points equal in number to our NeRF-based initialization method and apply multi-scale sampling. In the case of COLMAP, we utilize the entirety of the point cloud produced, given that the output size from COLMAP cannot be predetermined. 
Our NeRF-based initialization method demonstrates superior results in comparison to the other two methods. A visual comparison is provided in Fig.~\ref{fig:pointclouds}, where it is evident that grid-based NeRF (i.e., Instant-NGP~\cite{muller2022instant}) generates denser point clouds with improved geometric details and background modeling. 
This indicates that initializing with a grid-based NeRF captures the scene geometry more effectively.

\begin{table}
    \centering
    \caption{The ablation results of initialization, weighting strategies and color correction.}
    \resizebox{.58\columnwidth}{!}{%
    \begin{tabular}{@{}cc|cccl@{}}
    \toprule
    &           & PSNR$\uparrow$ & SSIM$\uparrow$ & LPIPS$\downarrow$ & FPS$\uparrow$ \\ \midrule
    \multirow{2}{*}{Init.}  & Random    &      21.20&      0.538&       0.513&     123.75\\
                            & COLMAP    &      23.00&      0.670&       0.365&     121.62\\
                            \midrule
    \multirow{3}{*}{Weighting}& Random    &      22.08&      0.635&       0.418&     122.84\\
                            & w/o weighting      &      22.50&      0.674&       0.357&     121.02\\
                            & Top1&      12.97&      0.241&       0.792&     \textbf{147.52}\\ 
                            \midrule
                            & w/o ColorCorr. &      23.86&      0.701&       0.328&     121.87\\ 
                            \midrule
                            & PyGS      &      \textbf{24.18}&      \textbf{0.728}&       \textbf{0.309}&     121.35\\ 
                            \bottomrule
    \end{tabular}%
    }
    \label{tab:ablation}
\end{table}

\noindent \textbf{Weighting Strategies:}
In Table~\ref{tab:ablation}, we compare our adaptive weighting strategy with three alternative approaches. The \textit{Random} strategy assigns a random weight to each pyramid level. The \textit{w/o weighting} strategy assigns a uniform weight of 1 to all levels. The \textit{Top1} strategy renders only the level with the highest weight. Our use of a learnable weighting network markedly enhances the rendering results compared to both the \textit{Random} and \textit{w/o weighting} approaches. Moreover, relying solely on the level with the maximum weight, as in the \textit{Top1} strategy, leads to instability during training and fails to achieve convergence.

\noindent \textbf{Effectiveness of Color Correction:}
We further investigate the impact of appearance embedding and color correction network in Table~\ref{tab:ablation}. Omitting the color correction network leads to a decline in performance metrics, underscoring the effectiveness of accounting for variations in lighting across different images.

\noindent \textbf{Impact of Pyramid Levels and Initialization Points:}
In Table~\ref{tab:pyramid-size}, we analyze the impact of the pyramid level $L$ and the number of initialization points $N$. Increasing both $L$ and $N$ results in improved rendering quality while causing a modest decrease in rendering speed.
This trend aligns with expectations, as additional pyramid levels enable the model to capture finer details of the scene, and the greater number of initialization points provides a more informed starting point for the model, offering an initial advantage for large scene modeling.

\begin{figure}[t]
    \centering
    \includegraphics[width=.98\linewidth]{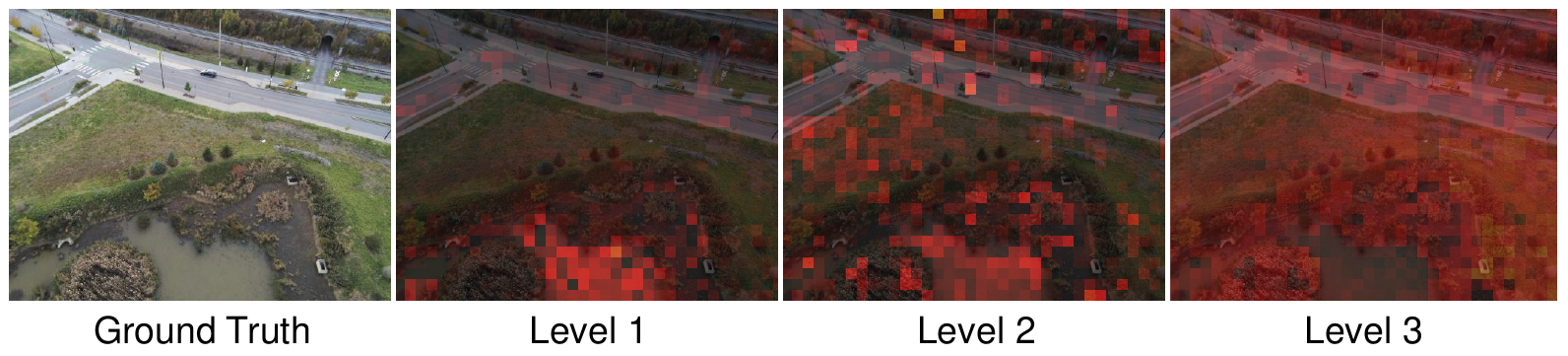}
    \vspace{-8pt}
    \caption{Visualization of weights of each level predicted by the weighting network.}
    \label{fig:weighting}
\end{figure}

\noindent \textbf{Behaviours of Weighting Network:}
In Fig.~\ref{fig:weighting}, we illustrate the distribution of weights assigned to each cluster across different pyramid levels upon projection onto the image plane. 
The visualization reveals that Gaussians at lower levels are particularly effective at capturing smooth objects that are nearer to the camera, such as the lake's surface. In contrast, Gaussians at the upper levels of the pyramid demonstrate a greater capacity for representing intricate and distant objects, such as the grass and trees.
\begin{table}[htp]
    \centering
    \vspace{-6pt}
	\begin{minipage}{0.38\textwidth}
		\centering
              \caption{The impact of number of pyramid levels and the size of initialization points.}
              \vspace{-6pt}
            \resizebox{0.95\textwidth}{!}{
            \begin{tabular}{@{}cc|cccc@{}}
            \toprule
            L                  & $N$ & PSNR$\uparrow$ & SSIM$\uparrow$ & LPIPS$\downarrow$ & FPS$\uparrow$ \\ \midrule
            \multirow{2}{*}{2} & 120k&      23.51&      0.686&       0.371&     159.6\\
                               & 240k&      23.81&      0.685&       0.345&     159.1\\ \midrule
            \multirow{2}{*}{3} & 260k&      23.97&      0.683&       0.333&     141.9\\
                               & 560k&      24.19&      0.728&       0.309&     124.2\\ \midrule
            \multirow{2}{*}{4} & 600k&      24.72&      0.734&       0.298&     117.3\\
                               & 1.2m&      24.93&      0.749&       0.287&     100.0\\ \bottomrule
            \end{tabular}%
            }
            \label{tab:pyramid-size}
        \end{minipage}
        \hfill
        \begin{minipage}{0.6\textwidth}
    	\centering
            \caption{Comparison of time usage of preprocessing, training and rendering between MegaNeRF~\cite{turki2022meganerf}, 3DGS~\cite{kerbl20233dgs} and our PyGS. PyGS takes significantly less time than other methods.}
            \vspace{-5pt}
            \resizebox{0.95\columnwidth}{!}{%
            \begin{tabular}{@{}c|ccc|ccc@{}}
            \toprule
                       & PSNR$\uparrow$ & SSIM$\uparrow$ & LPIPS$\downarrow$ & Pre.(h) & Train(h) & FPS\\ \midrule
            MegaNeRF   &      22.96&      0.588&       0.452&         -&          30:18&           0.01\\
            MegaNeRF-D &      22.34&      0.573&       0.464&         1:08&          30:18&           0.252\\
            3DGS-R     &      21.24&      0.586&       0.467&         -&          7:42&           128.70\\
            3DGS-C     &      22.52&      0.661&       0.354&         7:35&          11:53&           121.13\\ \midrule
            PyGS       &      24.19&      0.728&       0.309&         0:18&          13:30&           114.35\\ \bottomrule
            \end{tabular}%
            }
            \label{tab:time}
	\end{minipage}
\vspace{-16pt}
\end{table}

\noindent \textbf{Time Usage:}
In Table~\ref{tab:time}, we present a comparison of preprocessing, training, and rendering times across different methods, specifically among MegaNeRF~\cite{turki2022meganerf}, 3DGS~\cite{kerbl20233dgs}, and our proposal, PyGS.
MegaNeRF-D represents an accelerated version of MegaNeRF that distills the original MegaNeRF into an Octree-like architecture to improve rendering efficiency.
Table~\ref{tab:time} illustrates that both 3DGS-R and 3DGS-C achieve a significant speedup in training and rendering while maintaining rendering quality comparable to that of MegaNeRF and MegaNeRF-D. This demonstrates the substantial advantages of 3DGS-based methods over those based on NeRF.
Our approach, PyGS, outperforms the others in all three evaluated image-quality metrics (PSNR, SSIM, and LPIPS) while still offering time efficiency on par with the 3DGS methods. This underscores the efficacy of our method.
It is worth noting that, while 3DGS starting from COLMAP points has a marginally shorter training duration than PyGS, it requires several hours to generate point clouds using COLMAP. This additional step impacts the overall efficiency of 3DGS negatively.
\vspace{-10pt}
\section{Conclusion}
\vspace{-10pt}
In our study, we enhance the capabilities of the 3DGS framework \cite{kerbl20233dgs} without significantly increasing the training and rendering costs. 
We propose a novel pyramidal Gaussian architecture which initializes from a NeRF model. 
We perform cluster-level weighting by facilitating by a compact weighting network and learnable cluster embedding.
Our approach achieves notable advancements in representing large-scale scenes by effectively addressing inherent multi-scale challenges. However, the modeling of even larger environments requires substantial memory and computational resources. This indicates opportunities for future enhancements through the adoption of parallel optimization techniques.




\bibliography{iclr2024_conference}
\bibliographystyle{iclr2024_conference}


\end{document}